# Using Ensemble Models in the Histological Examination of Tissue Abnormalities

Giancarlo Crocetti, Michael Coakley, Phil Dressner, Wanda Kellum, Tamba Lamin
Seidenberg School of Computer Science and Information Systems
Pace University
White Plains, NY, USA
{gcrocetti, mcoakley, pd50340n, tl98810w, wk59882w}@pace.edu

*Abstract*—Classification models for the automatic detection of abnormalities on histological samples do exists, with an active debate on the cost associated with false negative diagnosis (underdiagnosis) and false positive diagnosis (overdiagnosis). Current models tend to underdiagnose, failing to recognize a potentially fatal disease.

The objective of this study is to investigate the possibility of automatically identifying abnormalities in tissue samples through the use of an ensemble model on data generated by histological examination and to minimize the number of false negative cases.

## I. INTRODUCTION

Breast cancer screening is conducted to detect cancerous cells before a person has symptoms. As part of a breast cancer prevention screening, if a lump is found, a fine-needle aspiration biopsy (FNAB) is performed, a technique which has been used widely in the evaluation of non-palpable breast lesions [1].

Typically, biopsy tissue samples from breast cancer are examined visually by a pathologist, who is looking for cancerous tissues with abnormal characteristics. However, the manual detection and quantification of such abnormality is still a tedious and laborious task. Today, automated image analysis systems can evaluate cytology features derived directly from a digital scan of breast FNAB slides [2].

Accuracy levels from manual analysis of samples has wide levels of accuracy ranging from 62.2% to 89.2% [1], while automatic methods based on three-factor Cox multivariate analysis [3] and clustering solutions [4] achieved much higher and consistent results with accuracies levels reaching 98%. Although such results are a definite improvement over manual diagnostic procedures, they still contain 2% of false negatives representing a failure of recognizing the associated sample as malignant, which can carry disastrous consequences.

In this study we are evaluating the performance of a model based on Classification and Regression Trees (CART) and Logistic Regression for the detection of malignancy in breast cancer tissues using an ensemble approach with the objective of reducing or eliminating the number of false negatives.

## II. THE DATA

In this study we utilized the Wisconsin Breast Cancer Dataset downloaded from the UC-Irvine machine learning archive [5] which contains 569 samples classified either as benign or malignant.

Each record consists of the following 12 attributes, containing tissue identification and outcome (attributes 1-2) and cellular characteristics (attributes 3-12):

1. Id
2. Diagnosis
3. Radius (mean of distances from center to points on the perimeter)
4. Texture (standard deviation of gray-scale values)
5. Perimeter
6. Area
7. Smoothness (local variation in radius lengths)
8. Compactness (perimeter$^2$ / area - 1.0)
9. Concavity (severity of concave portions of the contour)
10. Concave points (number of concave portions of the contour)
11. Symmetry
12. Fractal dimension (fdimension) ("coastline approximation" - 1)

The class to predict is "Diagnosis" and all attributes, with the exception of Id, will be considered as inputs. The attribute "Id" is excluded since it is used as a record ID and, therefore, completely unrelated to the experiment.

Out of the 569 records we generated:

a) A training set contains examination from 448 patients.
b) A test set contains examination from 121 patients.

## III. EXPLORATORY DATA ANALYSIS

### A. Outlier Detection

All the histological data is of good quality with the absence of missing values or outliers. For the detection of outliers we employed a Z-Score model which required the calculation of the following quantities:

$$x^* = \frac{x - \mu_x}{\sigma_x}$$

With this method we did identify few observations that fell outside the traditional [-4,+4] interval and, therefore, flagged as potential outliers [6], but after a careful review we decided to consider these values as valid, even though they were outside of the range.

*B. Normality Assumption*

Before proceeding with the modeling phase, we checked whether or not the variables satisfied the normality assumption, and indeed this was the case for all variables as shown in figure 1.

FIGURE 1. EXAMPLES OF DISTRIBUTION FOR THE RADIUS AND PERIMETER VARIABLES

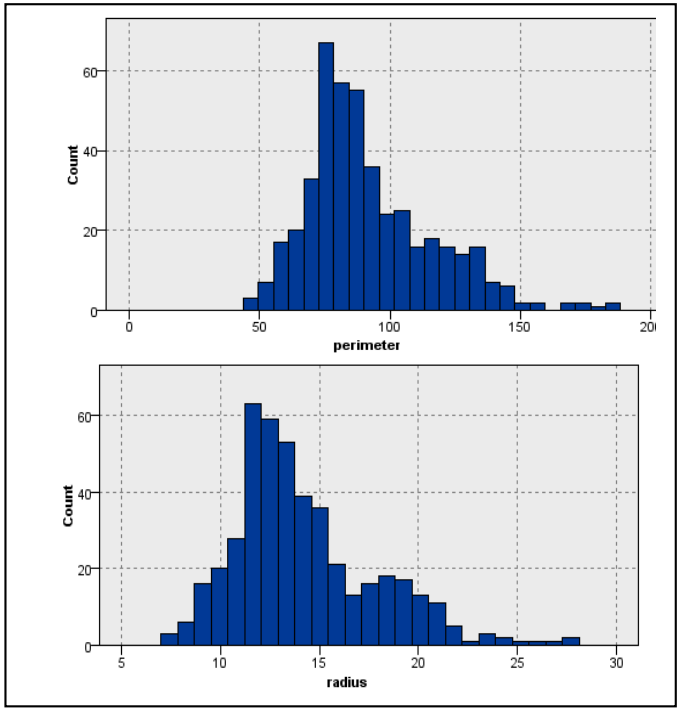

In order to further guarantee the quality of the data we calculated the Skewness and Kurtosis (i.e., peakness) measures for each variable and verified they were within the interval [-2/+2] considered the acceptable range for such statistics.

As reflected in their skewness levels, the variables displayed acceptable levels of symmetric; however, in terms of Kurtosis we had two variables that exceeded such range: area and fdimension, due to their long tails; a minor issue which was resolved with data normalization.

*C. Data Normalization*

Due to the large variations in the variables' range, we decided to normalize the data by applying a min-max transformation:

$$x^* = \frac{x - min_x}{max_x - min_x}$$

This transformation brought all the variables within the interval [0,1], guaranteeing that none of the variable will have higher influence due to their larger values [6].

*D. Correlation Analysis*

When calculating the Pearson correlation among the attributes, we found some strong correlation as shown in table 1.

TABLE 1 – PEARSON CORRELATION BETWEEN SOME ATTRIBUTES

| radius_mean_transformed | | |
|---|---|---|
| Pearson Correlations | | |
| perimeter_mean_transformed | 0.998 | Strong |
| area_mean_transformed | 0.994 | Strong |
| concavity_mean_transformed | 0.687 | Strong |
| concavepoints_mean_transformed | 0.820 | Strong |

Even though *concavity_mean_transformed* displayed a medium-high correlation with radius (0.687), it was strongly correlated with the other variables. Consequently, we decided to retain *radius_mean_transformed* and drop the other attributes.

## IV. CLUSTERING

With the use of the K-means clustering algorithm, we derived a new cluster attribute which divided the data in two cluster solution as shown in figure 2.

FIGURE 2. K-MEANS CLUSTER SOLUTION

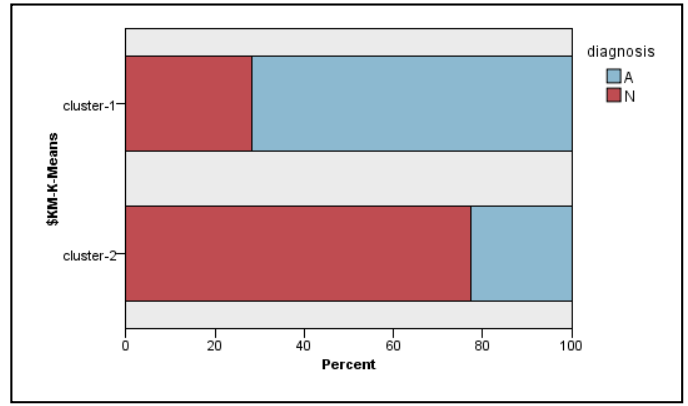

As shown in figure 2, cluster 2 contains a very high percentage of normal samples, an indication that this attribute might have some predictive power.

When considering which attribute played a more important role in generating this cluster solution we can see, from figure 3, that compactness, smoothness and symmetry played an important role which was also confirmed by comparing their cell distribution between clusters as shown in figure 4.

FIGURE 3. ATTRIBUTE IMPORTANCE

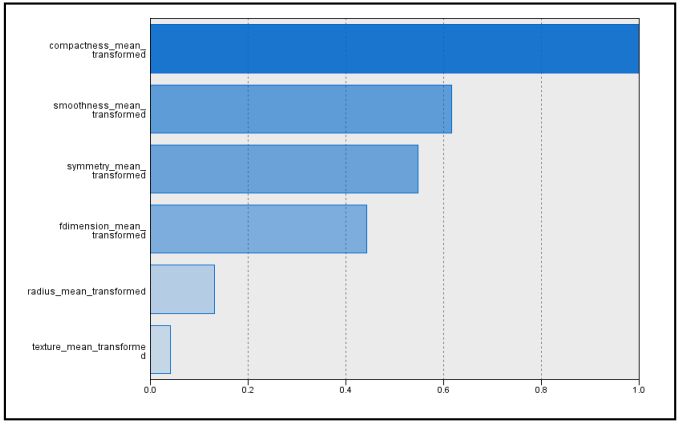

FIGURE 4. CELL DISTRIBUTION FOR COMPACTNESS, SMOOTHNESS, AND SYMMETRY

## V. MODELLING

Because of the continuous nature of attributes and the binary type of the targeted class, we decided to utilize the

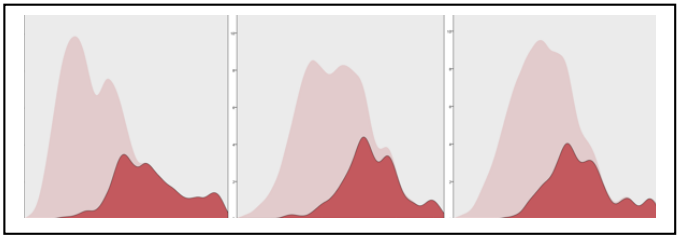

following models:

CART – Decision Tree
Logistic Regression

Each model performed quite well as we can see from the confusion matrixes in table 2 and the error rates in table 3

TABLE 2 – CONFUSION MATRICES FOR CART AND LOGISTIC REGRESSION

CART

| diagnosis | A | N |
|---|---|---|
| A | 40 | 1 |
| N | 12 | 68 |

Logistic Regression

| diagnosis | A | N |
|---|---|---|
| A | 39 | 2 |
| N | 4 | 76 |

TABLE 3 – ERROR RATES - PREDICTIVE MODELS

| Error | CART | Logistic Regression |
|---|---|---|
| Overall Error Rate | 0.10 | 0.05 |
| False Negative Rate | 0.02 | 0.03 |
| False Positive Rate | 0.23 | 0.09 |

The number of false positive is, however, quite large for CART, even though we also have 2 false negative in Logistic Regression which is more costly than any other misclassification type (failure in detecting a cancerous sample).

For the CART model is interested to take a look at the simple rules this model generated as in the following:

```
radius_mean_transformed <= 51.985 [Mode: N]
    compactness_mean_transformed <= 44.615 [Mode: N]
        radius_mean_transformed <= 45.475 [Mode: N] ⇒ N
        radius_mean_transformed > 45.475 [Mode: N]
            texture_mean_transformed <= 49.928 [Mode: N] ⇒ N
            texture_mean_transformed > 49.928 [Mode: A] ⇒ A
    compactness_mean_transformed > 44.615 [Mode: A] ⇒ A
radius_mean_transformed > 51.985 [Mode: A] ⇒ A
```

As we can see both radius and compactness play a very important role in the detection of abnormal tissues, even though, as mentioned earlier, the number of false positive is quite large.

In order to improve these results we thought of adopting an ensemble strategy by leveraging the confidence interval measures produced by these models.

Ensemble models have been considered an important development in Data Mining [7] and proven to improve model accuracy that is "easier and more powerful than judicious algorithm selection" [8].

In this particular we applied a voting scheme in which the prediction with the highest confidence wins.

When this ensemble model is put at work we were able to substantially improve the result as shown in table 4.

TABLE 4 – CONFUSION MATRIX FOR ENSEMBLE MODEL

| diagnosis | A | N |
|---|---|---|
| A | 40 | 1 |
| N | 4 | 76 |

Not only did we reduce considerably the total number of misclassifications, but we also improved the overall error rate associated to the predictive model, in fact:

$$Overall\ error\ rate = \frac{4+1}{121} = 0.04 = 4\%$$

$$False\ negative\ rate = \frac{1}{77} = 0.01 = 1\%$$

$$False\ positive\ rate = \frac{4}{44} = 0.9 = 9\%$$

These are quite good numbers even with a false positive rate close to 10%, representing cases that need to be reviewed in order to confirm the diagnosis.

## VI. CONCLUSIONS

The automatic classification of abnormal tissue samples is of paramount importance in helping physicians and other medical personnel in the diagnosis process.

The voting-based ensemble model derived through the combination of decision trees and logistic regression proved to be a very efficient way of helping in improving the detection of abnormal biopsy samples.

The very low false negative rate of 1% is a clear indication that this problem can be solved by the generation of high quality classification solutions, representing an improvement when compared to other classification systems developed in the past.